# Çok Alanlı Chatbot Mimarilerinde Avantajlı Performans ve Bellek Takası

D. Emre TAŞAR [1], Şükrü OZAN[2], M. Fatih AKCA[3], Oğuzhan ÖLMEZ[4],
Semih GÜLÜM[5], Seçilay KUTAL[5], Ceren BELHAN[6]

[1] Dokuz Eylül Üniversitesi, Türkiye, davutemre.tasar@ogr.deu.edu.tr
[2] AdresGezgini A.S , Ar-Ge Merkezi, Türkiye, sukruozan@adresgezgini.com
[3] Yönetim Bilişim Sistemleri, Sakarya Üniversitesi, Türkiye, mehmet.akca1@ogr.sakarya.edu.tr
[4] Yazılım Mühendisliği, Celal Bayar Üniversitesi, Türkiye, 172803041@ogr.cbu.edu.tr
[5] Mekatronik Mühendisliği, Marmara Üniversitesi, Türkiye, semihgulum@marun.edu.tr, secilaykutal@marun.edu.tr
[6] Yazılım Mühendisliği, İzmir Ekonomi Üniversitesi, Türkiye, ceren.belhan@std.ieu.edu.tr

**Özet**

Doğal dil işleme alanında metin sınıflandırma problemi oldukça geniş bir çalışma alanıdır. Metin sınıflandırma problemi kısaca, verilen metnin daha öncesinde belirlenen sınıflardan hangisine ait olduğunun tespit edilmesidir. Geçmiş çalışmalarda bu alan özelinde başarılı çalışmalar gerçekleştirilmiştir. Bu çalışmada doğal dil işleme alanında metin sınıflandırma probleminin çözümü için sıklıkla tercih edilen bir metot olan Dönüştürücüler için Çift Yönlü Kodlayıcı Temsilleri (BERT - Bidirectional Encoder Representations for Transformers) kullanılmıştır. Bir canlı sohbet botu mimarisinde kullanılmak üzere, farklı sınıflandırma problemlerini özel olarak eğitilmiş tek bir model üzerinden çözerek, her bir sınıflandırma problemi için ayrı bir model eğitilerek gerçekleştirilecek bir çözüm yöntemi ile sunucu üzerinde oluşacak işlemci, rastgele erişimli hafıza ve depolama alanı ihtiyacının hafifletilmesi amaçlanmaktadır. Bu noktada birden fazla konuda sınıflandırma yapılabilmesi için oluşturulan tek bir BERT modelinin tahmini sırasında uygulanan maskeleme yöntemi ile modelin tahmininin problem bazlı gerçekleştirebilmesi sağlanmıştır. Birbirinden farklı alanları kapsayan üç ayrı veri seti, problemin zorlaştırılması adına çeşitli yöntemlerle bölünmüş ve alan olarak birbirine çok yakın olan sınıflandırma problemlerine de bu sayede yer verilmiştir. Bu şekilde kullanılan veri seti 154 sınıfa sahip beş sınıflandırma probleminden oluşmaktadır. Tüm sınıflandırma problemlerini içeren bir BERT modeli ile problemler özelinde eğitilen diğer BERT modelleri performans ve sunucuda kapladıkları yer açısından birbirleriyle karşılaştırılmıştır.

*Anahtar Sözcükler: BERT; çoklu sınıf; doğal dil işleme; metin sınıflandırma.*

**Abstract**

Text classification problem is a very broad field of study in the field of natural language processing. In short, the text classification problem is to determine which of the previously determined classes the given text belongs to. Successful studies have been carried out in this field in the past studies. In the study, Bidirectional Encoder Representations for Transformers (BERT), which is a frequently preferred method for solving the classification problem in the field of natural language processing, is used. By solving classification problems through a single model to be used in a chatbot architecture, it is aimed to alleviate the load on the server that will be created by more than one model used for solving more than one classification problem. At this point, with the masking method applied during the estimation of a single BERT model, which was created for classification in more than one subject, the estimation of the model was provided on a problem-based basis. Three separate data sets covering different fields from each other are divided by various methods in order to complicate the problem, and classification problems that are very close to each other in terms of field are also included in this way. The dataset used in this way consists of five classification problems with 154 classes. A BERT model containing all classification problems and other BERT models trained specifically for the problems were compared with each other in terms of performance and the space they occupied on the server.

*Keywords: BERT; multi-class; natural language processing; text classification.*

## 1. Giriş

Yapay zekanın alt dallarından biri olarak kabul edilen Doğal Dil İşleme; metin verileri üzerinde duygu sınıflandırma, özet bilgi çıkarma, metin sınıflandırılması gibi görevlerle ilgilenen bir alandır [1]. Günümüzde teknolojinin gelişmesi sayesinde doğal dil işleme alanında da gelişmeler meydana gelmektedir. Doğal dil işleme kullanılarak üretilen canlı sohbet botları teknolojinin de gelişmesiyle insansı bir davranış sergileyebilir hale gelmiştir.

Muangkammuen vd. tarafından yapılan çalışmada e-ticaret sektöründe sıkça sorulan soruları cevaplamak için LSTM tabanlı bir canlı sohbet botu modeli oluşturulmuştur. Oluşturulan LSTM modelinin doğruluk değeri %83 olarak hesaplanmıştır [2].

Ozan vd. gerçekleştirdikleri çalışmada çeşitli doğal dil işleme modellerinin kategorik sınıflandırma problemi üzerindeki performanslarını karşılaştırmışlardır. LSTM, BERT ve Doc2Vec modellerinin karşılaştırıldığı çalışmada en başarılı modelin %93 doğruluk seviyesiyle BERT modeli olduğu sonucuna varılmıştır [3]. Bu nedenle bu çalışma da BERT modeli üzerinden gerçekleştirilmiştir.

Tunç vd. tarafından yapılan çalışmada 16 kategorili bir veri setinde BERT modeli ile metin sınıflandırma yapılmıştır. Çalışmada %90 doğruluk değeri hedeflenmiş ancak %79 doğruluk değerine ulaşılmıştır. Modelin istenilen doğruluk seviyesine ulaşamamasının veri setinde yer alan cümlelerin, otomatik etiketleme yapılabilmesi için içerisinde net bilgi içermemesinden kaynaklı olduğu sonucuna varılmıştır [4].

Bahsedilen çalışmalar doğrultusunda gerçekleştirilen bu çalışmada ise, bir canlı sohbet botu mimarisinde kullanılmak üzere gelen isteğe göre önceden hazırlanmış yanıtlar arasından en uygununu seçebilen bir sistem kurgulanmıştır. Kurgulanan sistemde derin öğrenme metodolojisi olarak güncel teknolojiler arasında yer alan BERT modeli tercih edilmiştir [3].

Geliştirilen canlı sohbet botu mimarilerinde farklı firmalar ve projeler için oluşturulacak olan her bir model bellekte ayrı bir yer işgal etmektedir. Çözümü tek bir modele indirgeyebilmenin farklı firmalara hizmet veren şirketler için sistem kaynağı gereksinimi açısından etkili bir optimizasyon yöntemi olabileceği öngörülmektedir.

Öncelikle beş farklı veri seti kendi özelindeki BERT modelinin eğitimi ile gerçekleştirilen sınıflandırma işlemlerinin başarımları hesaplanmıştır. Daha sonra, bütün veri setlerinin birleştirilmesiyle oluşan toplamda 154 sınıflı bir veriyle tek bir BERT modelinin eğitilmesinin ardından gerçekleştirilen sınıflandırma işleminin başarımları hesaplanarak, elde edilen sonuçlar karşılaştırılmıştır. Bu sayede sunucuda tek bir model çalıştırılması ile daha az maliyetli bir BERT modelinin, ayrı ayrı ilgili konular özelinde eğitilen modellere yakın düzeyde başarılı olabileceği gösterilmiştir.

## 2. Materyal ve Yöntem

### 2.1. Kullanılan veri seti

Çalışmada "banka", "firma" ve "covid" şeklinde adlandırılacak olan üç ayrı veri setine yer verilmiş, bu veri setleri çeşitli yöntemlerle bölünerek veri seti sayısı beşe çıkarılmıştır. Böylelikle tüm veri setlerini içeren modelin eğitimi sırasında benzer cümleler içeren alt veri setlerinin sınıflandırma esnasında birbiriyle karışma problemine de çalışmada yer verilmiştir.

Banka isimli veri seti, 77 sınıfa sahip, bir bankaya müşterileri tarafından yöneltilen sorular ve soruların kategori isimlerinden oluşmaktadır [5]. Huggingface veri setleri kütüphanesinden alınan bu veri seti öncelikle makine çevirisi ile Türkçe'ye çevrilmiş ve çevrilen cümlelerin çeviri kalitesi manuel olarak kontrol edilmiştir. Bu veri seti kategori sayısına göre ikiye bölünerek 38 ve 39 etikete sahip iki ayrı alt veri kümesi (Veri 1 ve Veri 2) elde edilmiştir.

Birbiriyle aynı veriler içeren iki ayrı veri seti elde edilmesi amacıyla Firma isimli 26 sınıflık veri setinde bulunan her bir sınıf, içerdiği veri sayılarına göre ikiye bölünmüştür. Böylelikle bu veri setinden 26 sınıfa sahip iki ayrı alt veri kümesi (Veri 3 ve Veri 4) elde edilmiştir.

Covid adlı veri seti (Veri 5) ise Covid-19 karantina yasakları kapsamında bakanlığa yöneltilen sorulardan ve soruların kapsadığı alanlardan oluşmaktadır. Bu veri seti T.C. İçişleri Bakanlığı'nın internet sitesi üzerindeki sıkça sorulan sorular ve yanıtları baz alınarak manuel olarak üretilmiştir.

Veri setinin tamamı 3-113 arasında kelime sayısına sahip uzun ve kısa cümlelerden oluşurken cümle uzunluklarına göre düzensiz bir dağılım sergilemektedir. Verinin sahip olduğu beş adet alt küme veri seti, içerdiği etiket sayısına göre ve her bir etiketin sahip olduğu veri sayısına göre de düzensiz dağılıma sahiptir.

**2.2. BERT yöntemi**

Makalede sınıflandırma problemlerinin çözümü için Devlin vd. tarafından geliştirilen BERT modeli [6] kullanılmıştır. BERT yöntemi 12 dönüştürücü bloğu, 12 öz-dikkat başlığı ve 768 gizli katmandan oluşan çift yönlü bir dönüştürücü modelidir. Bu modelin eğitimi iki aşamadan meydana gelir. İlk aşama, yüksek boyutta veriler ile gerçekleştirilen ön eğitim aşamasıdır. Modelin bu aşamada dil yapısını kavraması beklenmektedir. İkinci aşama olan ince ayar ise problem bazlı kullanılacak veri seti ile modelin yeniden eğitilmesi işlemidir. Bu işlem ile model, probleme özgü bir dikkat mekanizması oluşturur ve probleme yönelik tahminlerini güçlendirir.

İnce ayar işlemi için seçilen ön eğitimli model, kategorik sınıflandırmada en yüksek başarım elde edilen [7] lodoos/bert-base-turkish-cased [8] modelidir. Bahsedilen model 200 GB boyutunda çeşitli Türkçe metinlerden oluşan bir veri setiyle ön eğitime tabi tutulmuş bir model olup, çalışmada beş veri seti ve ek olarak bu veri setlerinin tümünün birleşiminden oluşan toplamda altı veri seti, beş dönem (epoch) süresince eğitilerek ince ayar işlemi toplam altı kere gerçekleştirilmiştir.

Model başarımlarının değerlendirilmesi için çeşitli kriterler ele alınmıştır. Çoklu sınıflandırma problemlerinin değerlendirilmesi için genel olarak kullanılan doğruluk değeri ve hata düzeyinin (confusion matrix) yanı sıra sıklıkla tercih edilen F1 skoru ve makro F1 skoru kriterleri de göz önüne alınmıştır.

**2.3. Tahminleme fonksiyonu ve alan tasarrufu**

Çalışmada kullanılan BERT modeli, çok kategorili sınıflandırma görevinde ince ayara tabi tutulduğu için, kategori sayısı kadar etiket ile sınıflandırma modelini etiketleyebilme becerisine sahiptir. Fakat bu durumda BERT modelinin, farklı unsurlara hizmet edecek tek bir chatbot için üretilmiş olan bellek verimli BERT modelinde sorguya en yakın etiketleme işlemini ait olmadığı bir unsura yönlendirmesi sonucu modelin doğruluk değerlerinin %52'ye kadar düştüğü görülmüştür. Bu sorunu gidermek için BERT modelinin tahminlemesini, sadece o unsura ait olarak modelde bulunan kategoriler arasında yapmasını sağlayacak şekilde kısıtlayan bir fonksiyon yazılmıştır. Bu tahminleme kısıtlama fonksiyonu sayesinde modelin, her unsur özelinde başarı oranları tekrar %90'ın üzerine yükselmiştir.

**3. Bulgular**

İnce ayar yapılan toplamda 6 model üzerinden, oluşturulan tahminleme fonksiyonu yardımıyla tahmin yapılmış, sonuçları değerlendirmek için kayıp, doğruluk ve F1 skoru metriklerinden yararlanılmıştır. Doğruluk ve F1 skoru, eğitim ve test veri seti için ayrı ayrı hesaplanmıştır. Model dönem grafikleri Şekil 1'de paylaşılmıştır.

Modelin tahmini esnasında veri setlerine dair ön bilgi içeren ve bu sayede veri setine özgü bir tahmin yürütme imkanı sunan tahminleme fonksiyonunun modele katkısı incelenmiştir. Tüm veri seti ile eğitilmiş model, tahminleme fonksiyonu kullanılarak test edilmesinin yanı sıra ayrıca tahminleme fonksiyonu kullanılmadan da aynı test verisi üzerinden bir teste tabi tutulmuştur. Tahminleme fonksiyonu kullanılmadan gerçekleştirilen test ile elde edilen %52 doğruluk seviyesi, tahminleme fonksiyonu kullanıldığında %90 doğruluk seviyesine çıkmıştır. Her iki doğruluk değeri kıyaslandığında modelin tahmin fonksiyonu sayesinde başarı konusunda oldukça gelişim gösterdiği sonucuna varılabilmektedir.

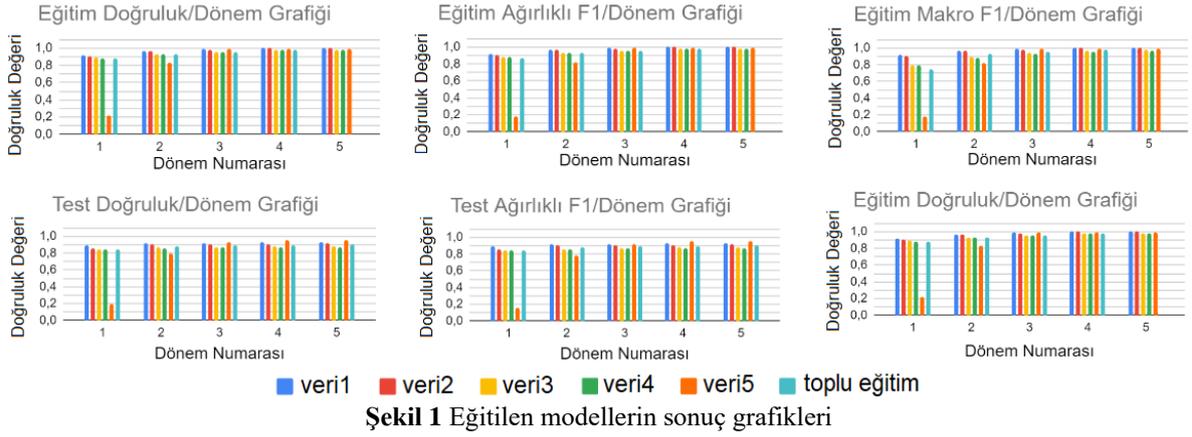

Şekil 1 Eğitilen modellerin sonuç grafikleri

## 4. Sonuçlar ve Öneriler

Çalışmada, farklı firma verileri içeren ve tahmin yaptırılacak firmaya özgü sınıf aralıklarına dair ön bilgiye sahip tek bir BERT modelinin, her bir firma için o firmaya ait veri setiyle ince ayar yapılan diğer BERT modelleriyle başarımı kıyaslanmış ve bulunan çözüm ile bellek alanından büyük oranda tasarruf sağlanması hedeflenmiştir. Modelin eğitimi esnasında birbirine çok yakın konularda sınıflandırma problemlerine (birbirinden ayrı veri setleri) yer verilerek problem kasıtlı olarak zorlaştırılmıştır.

Veri setinin tamamıyla eğitilmiş modelin sınıf tahmini esnasında doğruluğun arttırılması için direkt olarak 154 sınıf arasından tahmin edilmesi yerine girdinin ait olduğu alt veri setinin sınıf aralıklarına dair ön bilgi verilmektedir. Bu sayede yalnızca veri setinin sahip olduğu sınıf aralığında tahmin yapılır. Kullanılan beş veri seti ile ayrı ayrı yapılan eğitimlerde test veri setleri ile ortalama %91,2 doğruluk değerine ulaşılmıştır. Buna ek olarak birçok sınıflandırma problemi içeren tek BERT modeli, veri setine ait ön bilgiye sahip olup olmamasına göre tahmin yapması sağlanarak sonuçlar incelenmiştir. Kullanılan alt veri setlerinin (firmaların) kapsadığı sınıflara dair bir ön bilgi içeren tahminleme fonksiyonu kullanılmadan test edilen modelde %52 doğruluk seviyesi elde edilmişken, tahminleme fonksiyonu kullanılarak %90 doğruluk değerine ulaşılmıştır. Başarım oranları arasındaki büyük farkın asıl sebebi modelin hangi aralıktan tahmin yapacağı ön bilgisine sahip olmasıdır.

Kullanılan ön eğitimli BERT modeline göre ince ayar edilmiş olan bir model bellekte 1.6 GB yer tutmaktadır. Bu durum göz önüne alındığında her bir veri seti özelinde eğitilmiş beş adet model 8 GB yer kaplarken, beş veri setiyle eğitilen tek bir model yalnızca tek bir model kadar (1.6 GB) yer kaplamaktadır. Sonuçlar göz önüne alındığında veri seti özelinde eğitilmiş çok sayıda modelden, kullanılan veri setlerinin tamamıyla eğitilmiş tek modele geçişte %1,2'lik doğruluk değeri kaybı olduğu gözlemlenirken oluşturulan yapı sayesinde bellek alanından yaklaşık %80 tasarruf edilmiştir.